\documentclass[10pt,twocolumn,letterpaper]{article}

\usepackage[pagenumbers]{wacv} 

\usepackage{graphicx}
\usepackage{amsmath}
\usepackage{amssymb}
\usepackage{booktabs}
\usepackage{multirow}
\usepackage{enumitem}
\usepackage[accsupp]{axessibility}  

\usepackage[pagebackref,breaklinks,colorlinks]{hyperref}

\usepackage[capitalize]{cleveref}
\crefname{section}{Sec.}{Secs.}
\Crefname{section}{Section}{Sections}
\Crefname{table}{Table}{Tables}
\crefname{table}{Tab.}{Tabs.}

\def\wacvPaperID{1373} 
\def\confName{WACV}
\def\confYear{2024}

\begin{document}

\title{Movie Genre Classification by Language Augmentation and Shot Sampling}

\author{Zhongping Zhang$^{1}$ \qquad Yiwen Gu$^{1}$ \qquad Bryan A. Plummer$^{1}$ \\
\qquad Xin Miao$^{2}$ \qquad Jiayi Liu$^{2}$ \qquad Huayan Wang$^{2}$ \\
$^{1}$Boston University \qquad $^{2}$Kuaishou Technology \\
{\small $^1$\texttt{\{zpzhang, yiweng, bplum\}@bu.edu} \qquad $^2$\texttt{wanghy514@gmail.com}}
}
\maketitle

\begin{abstract}
    Video-based movie genre classification has garnered considerable attention due to its various applications in recommendation systems. Prior work has typically addressed this task by adapting models from traditional video classification tasks, such as action recognition or event detection. However, these models often neglect language elements (\eg, narrations or conversations) present in videos, which can implicitly convey high-level semantics of movie genres, like storylines or background context. Additionally, existing approaches are primarily designed to encode the entire content of the input video, leading to inefficiencies in predicting movie genres. Movie genre prediction may require only a few shots\footnote{A shot is defined as a series of frames captured from the same camera over an uninterrupted period of time~\cite{thirard1994robert}.} to accurately determine the genres, rendering a comprehensive understanding of the entire video unnecessary.  To address these challenges, we propose a Movie genre Classification method based on Language augmentatIon and shot samPling (Movie-CLIP). Movie-CLIP mainly consists of two parts: a language augmentation module to recognize language elements from the input audio, and a shot sampling module to select representative shots from the entire video. We evaluate our method on MovieNet and Condensed Movies datasets, achieving approximate $6-9$\% improvement in mean Average Precision (mAP) over the baselines. We also generalize Movie-CLIP to the scene boundary detection task, achieving $1.1$\% improvement in Average Precision (AP) over the state-of-the-art. We release our implementation at \href{https://github.com/Zhongping-Zhang/Movie-CLIP}{this http URL}.

\end{abstract}

\section{Introduction}
\label{sec:introduction}

\begin{figure}[t]
    \centering
    \includegraphics[width=1.0\linewidth]{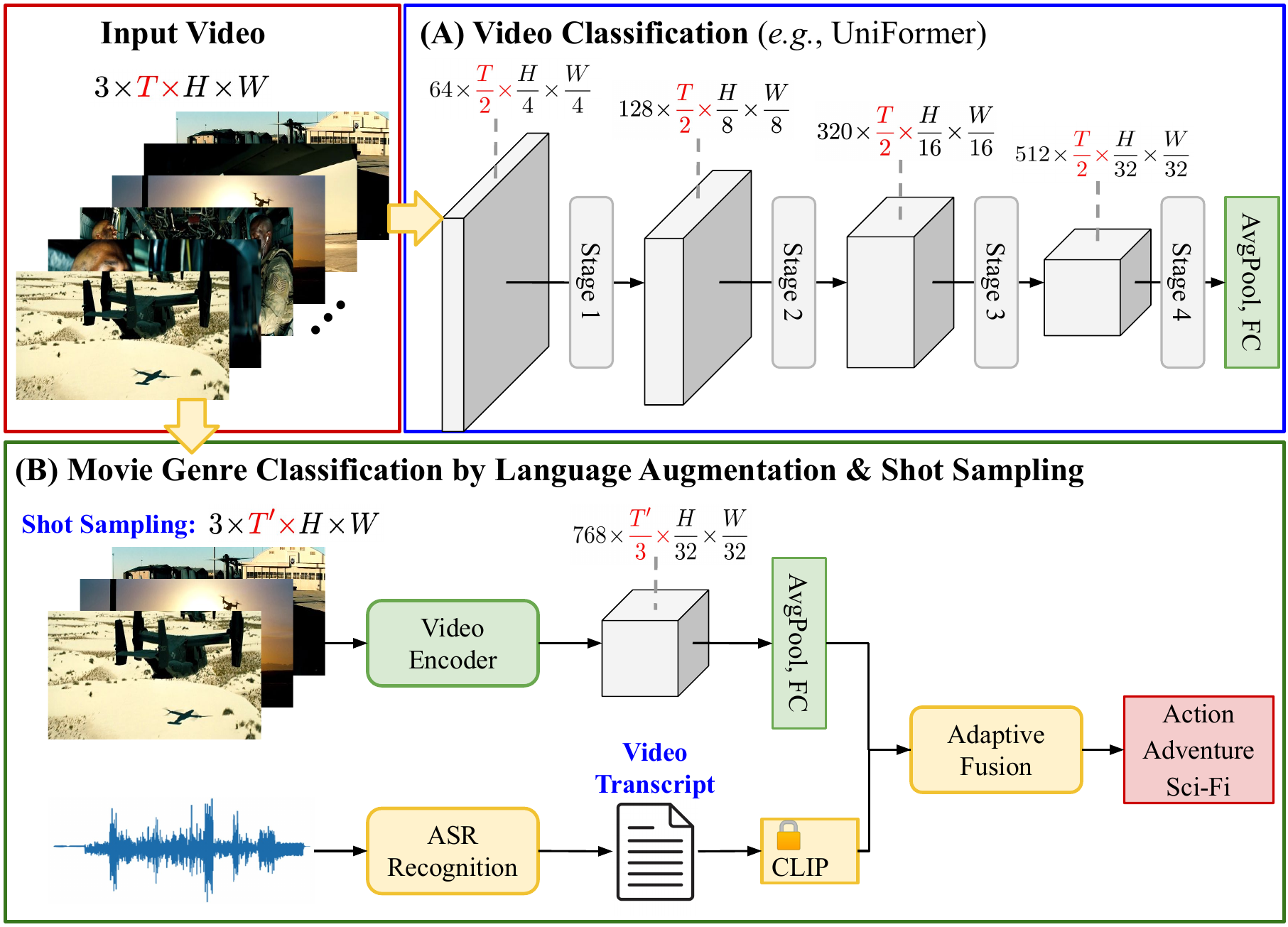}
    \vspace{-2mm}
    \caption{Our task aims at predicting movie genres based on input videos. Existing methods commonly approach this task by adapting models from related video-classification tasks (\eg, action recognition~\cite{li2023uniformer,carreira2017quo} or topic recognition~\cite{abu2016youtube}). As shown in (A), these methods typically incorporate the entire video as input and ignore the language elements in videos, making the prediction less efficient and accurate. To address these challenges, we introduce two components to our model, as shown in (B). We propose a language augmentation module to extract language information from the input video, circumventing the dependence on provided video captions and improving the prediction accuracy. Additionally, we apply shot sampling strategy to select representative shots from the entire video, leading to a notable reduction in computational cost.
    }
    \vspace{-4mm}
    \label{fig:overview}
\end{figure}

Video-based movie genre classification facilitates a wide range of applications, including content recommendation~\cite{deldjoo2016content,Lee_2017_ICCV}, genre-based video retrieval and filtering~\cite{huang2020movienet}, and automatic tagging and annotation~\cite{huang2018trailers}. Early research on this task typically focused on domain-specific datasets with only a few categories~\cite{rasheed2005use, brezeale2006using, zhou2010movie, simoes2016movie}. Restricted by the model's scalability and dataset size, these methods primarily predict movie genres using posters or still frames from videos. Recently, to leverage the prior knowledge of large-scale video-based datasets~\cite{abu2016youtube, carreira2017quo, huang2020movienet}, researchers have explored adapting models from related video classification tasks (\eg, Uniformer~\cite{li2023uniformer} or TRN~\cite{zhou2018temporal}) to movie genre classification, as shown in \Cref{fig:overview} (A).

Directly adapting these video classification models to movie genre prediction presents two challenges. First, prior work typically neglects the language information present in videos~\cite{carreira2017quo, zhou2018temporal, feichtenhofer2019slowfast, xiao2020audiovisual, li2023uniformer}. However, this information, such as narrations or conversations, can implicitly convey genre information. In some cases, movie genres can even be accurately predicted based solely on movie transcripts. For example, imagine a movie scene where characters engage in intense discussions about mysteries, accompanied by eerie music. The suspenseful dialogue and atmospheric setup convey the information of horror or thriller genre. In contrast, in a movie where characters engage in coincidental encounters and comedic misunderstandings, the lighthearted language will suggest that the film most likely belongs to the romantic or comedy genre. Second, video classification frameworks often encode the entire video to comprehend the events occurring in videos. A comprehensive understanding of the entire video is essential for tasks like action recognition or event detection. However, in movie genre classification, we observe that only a few shots can be sufficient to accurately predict movie genres. For example, humans can predict genres based solely on clips or trailers, without the need to watch the entire movie.

To introduce language information into models, a straightforward method is collecting text documents related to the input video and encoding them as part of the input. This strategy is often adapted by multimodal methods when provided with both videos and text documents~\cite{cascante2019moviescope, mangolin2022multimodal, cai2023multi}. However, collecting the text documents will introduce extra overhead, and these text documents may not always be available. For example, Condensed Movies~\cite{bain2020condensed} collected approximately 33,000 movie clips from YouTube, and captions were absent in half of these videos. To address this issue, we integrate an Automatic Speech Recognition (ASR) system, Whisper~\cite{radford2023robust}, into our model, as illustrated in \Cref{fig:overview} (B). In this scenario, our model automatically extracts language information from the audio, eliminating the requirement of user-provided captions. Additionally, we introduce a shot sampling module to improve the efficiency of our model. Motivated by sparse sampling strategies~\cite{wang2016temporal, huang2018trailers, lei2021less}, our method improves efficiency from two aspects. First, we segment the entire video into individual shots and select shots from different scenes as the video representation. Second, we subsample keyframes from each shot to use as the shot representation. Since the number of keyframes $T'$ is significantly smaller than the total number of frames $T$ in the video (\Cref{fig:overview} (B)), our model notably reduces the computational cost.

In summary, the contributions of this paper are: 
\begin{itemize}[nosep,leftmargin=*]
  \item We propose a language augmented approach for movie genre prediction. Compared to prior work~\cite{huang2018trailers, huang2020movienet, cascante2019moviescope}, Movie-CLIP extracts language information from the input video and does not rely on external language sources, such as captions, metadata, or Wikipedia. 
  \item We leverage a shot sampling strategy to select key frames from input video as the visual representations. This strategy notably reduces the computational cost, while achieving competitive performance compared to frameworks~\cite{zhou2018temporal, li2023uniformer} that encode the entire video.
  \item Experiments on MovieNet~\cite{huang2020movienet} and Condensed Movies \cite{bain2020condensed} demonstrate that Movie-CLIP outperforms the baselines, improving approximately 6-9\% mAP points on genre classification. We further show that Movie-CLIP generalizes to scene boundary detection task, achieving 1.1\% improvement in AP over the state-of-the-art.
  \item We perform extensive experiments on movie genre classification, exploring the correlations between movie genres and different components across various modalities.
\end{itemize}

\section{Related Work}
\noindent \textbf{Studies on Movies} involve a great number of research topics, spanning genre classification~\cite{huang2018trailers, huang2020movienet, cascante2019moviescope}, scene boundary detection~\cite{rao2020local, chen2021shot}, shot boundary detection~\cite{souvcek2019transnet, souvcek2020transnet}, person re-identification~\cite{xia2020online}, action recognition~\cite{bojanowski2013finding, sigurdsson2016hollywood, xu2021long}, alignment between movie and text descriptions~\cite{zhu2015aligning, tapaswi2015book2movie, cour2008movie}, understanding relationships of film characters~\cite{park2009character, weng2009rolenet, bamman2013learning, kukleva2020learning}, movie question answering~\cite{tapaswi2016movieqa, wang2018movie, kim2019progressive}, scene and event understanding~\cite{chen2021shot, sadhu2021visual}, among others. Existing approaches typically comprehend movies from a visual perspective~\cite{huang2018trailers,huang2020movienet} or align the visual modality with corresponding labels across other modalities, such as actions~\cite{kuehne2011hmdb} or text descriptions~\cite{tapaswi2015book2movie}. Therefore, the language elements in movies are either ignored or provided as part of the input. In this paper, we explore to automatically extract language elements from input videos to improve the performance of genre classification. Compared to previous multimodal methods (\eg, Moviescope~\cite{cascante2019moviescope}), Movie-CLIP leverages the language information for free, eliminating the requirement of additional language annotations like Wikipedia or metadata.
\smallskip

\noindent \textbf{Movie Genre Classification} can be divided into two major categories: image-based (posters, still frames, etc.)~\cite{zhou2010movie, simoes2016movie, huang2020movienet} or video-based (trailers, movie clips, etc.)~\cite{huang2018trailers, huang2020movienet}. Recently, researchers have adapted popular frameworks from traditional video classification tasks to movie genre classification, such as methods on action recognition~\cite{tran2015learning, wang2016temporal, zhou2018temporal, meng2020ar, chen2021deep} or video summarization~\cite{potapov2014category, truong2007video}. An obstacle for these frameworks is the computational cost. Methods that take all frames as input would be infeasible to handle videos with hours' duration~\cite{tran2015learning, wang2016temporal}. Though sparse sampling strategies have been proposed to process videos more efficiently~\cite{zhou2018temporal, lei2021less}, the analysis of hour-long videos still costs significant resources. To address this issue, we use a shot sampling algorithm to first divide the entire video into individual shots, and then select representative shots from different scenes, predicting genres efficiently.
\smallskip

\noindent \textbf{Scene Boundary Detection} aims to identify the starting and ending points of different scenes in videos. Early methods~\cite{rasheed2003scene, rui1998exploring} primarily employed unsupervised learning to segment scenes, relying on the similarity in colors. With the emergence of datasets with human-annotations~\cite{rotman2017optimal, baraldi2015deep}, supervised learning approaches~\cite{baraldi2015deep, protasov2018using, rao2020local, rotman2017optimal} have been proposed. A notable advancement in this topic was marked by the introduction of MovieNet~\cite{huang2020movienet}, which comprises 1,100 movies, with 318 of them annotated with scene boundaries.

\begin{figure*}[t]
    \centering
    \includegraphics[width=\linewidth]{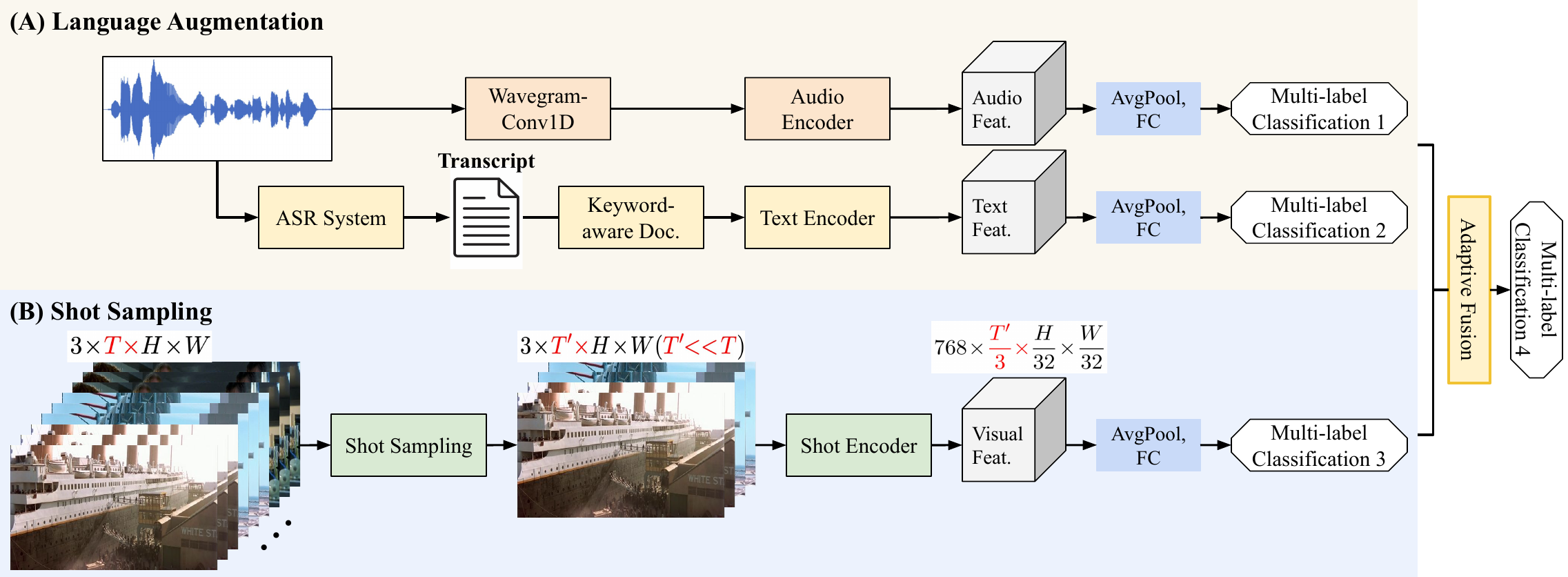}
    \vspace{-1mm}
    \caption{\textbf{Movie-CLIP Overview}. Our approach consists of two major components: (A) Language Augmentation: An ASR System, Whisper~\cite{radford2023robust}, is employed to automatically extract language elements for the input video, eliminating the requirement of provided captions. We further propose a keyword-aware mechanism to suppress the noise introduced by the ASR system. See Section \ref{sec:movieclip_language} for detailed disccusion; (B) Shot Sampling: We introduce a sparse sampling strategy to select shots from different scenes as the video representation. In each shot, we sample key frames as the shot representation. See Section \ref{sec:movieclip_shotsampling} for detailed discussion.
    }
    \vspace{-2mm}
    \label{fig:MovieCLIP_framework}
\end{figure*}

\section{Movie-CLIP: Movie genre Classification by Language augmentatIon \& shot samPling}
\label{sec:movieclip}

Given a user-provided video $V$, our task aims at predicting movie genres accurately and efficiently by leveraging language elements $L$ and sparsely sampled video representation $V'$. Since the language elements (\eg, narrations or conversations) in videos can convey high-level semantics of movie genres, we explore the incorporation of language information, such as movie transcripts, into genre classification models. To circumvent the dependence on user-provided text documents, we integrate an ASR system to automatically recognize language elements from the audio, as discussed in \Cref{sec:movieclip_language}. To improve model efficiency, we apply a shot sampling module in \Cref{sec:movieclip_shotsampling}. This module selects representative shots from the entire video as the visual representation. Additionally, we introduce a fusion strategy to concatenate the outputs of each modality, which is discussed in \Cref{sec:movieclip_fusion}. \Cref{fig:MovieCLIP_framework} provides an overview of our method.

\subsection{Language Augmentation}
\label{sec:movieclip_language}

As discussed in the Introduction, language information such as narrations or conversations can play an important role in genre prediction. For example, narrations of \emph{documentary} genres are often richer than those in \emph{action} genres. Existing multimodal approaches commonly incorporate language information from user-provided text documents~\cite{cascante2019moviescope, cai2023multi}. However, while some videos come with text data like captions, there is a considerable number of videos that do not have captions. To circumvent the dependence on provided text documents, we introduce a language augmentation module in Movie-CLIP. This module incorporates an ASR model, Whisper~\cite{radford2023robust}, to generate transcripts. In other words, the input to Movie-CLIP comprises the input video and its associated audio, from which the language modality is extracted and leveraged by our model without additional requirements. Thus, we named this mechanism as the language augmentation module. An overview of our language augmentation module is provided in \Cref{fig:MovieCLIP_framework} (A), consisting of the following components:
\smallskip

\noindent \textbf{Automatic Speech Recognition.} Given the input audio $A$, we apply Whisper~\cite{radford2023robust} to obtain the initial transcript $L$. The initial transcript consists of multiple language components, including narrations, conversations, voice activities, etc.
\smallskip

\noindent \textbf{Keyword-aware Documents.} To incorporate the language information, a straightforward method is to directly apply a language encoder to the transcript $L$. However, in our experiments, we observed that directly applying a language encoder like BERT~\cite{kenton2019bert} to $L$ only resulted in a slight boost to the performance of our genre prediction model. We attribute this to the fact that the ASR system cannot perfectly recognize all language tokens, leading to the presence of noise in the initial transcript $L$, which might adversely affect the genre prediction results. In addition, as we will show, some words are predictive of a particular genre. Thus, we propose a keyword-aware algorithm based on these observations. Motivated by the intuition that Nouns, Pronouns, and Adjectives often contain important clues for describing events in videos, we first narrow the scope from the entire document to Nouns, Pronouns, and Adjectives. Each word's part-of-speech is identified by SpaCy~\cite{honnibal2017spacy}. We then define keyword $K$ as tokens with high frequency that appear in captions. From these tokens, we select the top $k$\footnote{In our experiments, we set $k$ to 20.} tokens with high frequency that appear in captions. 
\smallskip

\noindent \textbf{Language Representation.} Our language encoder takes the concatenation of the initial transcript $L$ and the keywords $K$ as input. Language representations are obtained by ${\rm f}_l(L,K; \boldsymbol{\theta_l})$, where we apply the text encoder of CLIP~\cite{radford2021learning} as our $f_l(\cdot)$. Based on ${\rm f}_l(L,K; \boldsymbol{\theta_l})$, we further apply a linear layer to get the prediction score for the language modality: 
\begin{align}
    \rho_{l} = W{\rm f}_l(L,K; \boldsymbol{\theta_l})+b,
\end{align}
where $W$ and $b$ are the linear layer's learnable parameters.
\smallskip

\noindent \textbf{Language Modality Loss.} As shown in \Cref{fig:MovieCLIP_framework}, we apply a multi-label classification head\footnote{Since a movie often has multiple genre labels, movie genre prediction is formulated as a multi-label classification task.} to each modality, enabling our model to effectively handle both multi-modal features and each individual modality. We apply binary relevance to train our model. The language prediction head of Movie-CLIP comprises an ensemble of binary classifiers, with each classifier predicting the presence of a specific genre in the video. The loss of the $j$-th genre can be expressed as
\begin{align}
    L_{lj}=- \left[ y_{j}\log(\rho_{lj})+(1-y_{j})\log(1-\rho_{lj}) \right],
    \label{equation:loss_language}
\end{align}
where $y_j$ and $\rho_{lj}$ denote the ground truth label and prediction scores for the $j$-th movie genre, respectively. We average Eq.~\ref{equation:loss_language} across each classifier to obtain the final text loss: 
\begin{align}
    L_{text} = \frac{1}{K}\sum^{K}_{j=1} L_{lj}.
\end{align}

\noindent \textbf{Audio Representation and Loss.} Consistent with the language representation and loss, we obtain the prediction score of the audio modality, $\rho_a$, by applying PANNs~\cite{kong2020panns}, and use a binary relevance strategy to train the multi-label classifiers for the audio modality.

\subsection{Shot Sampling}
\label{sec:movieclip_shotsampling}
In movie genre classification, the input video can be quite lengthy in some cases (\eg, a movie may last around two hours or longer). Consequently, encoding the frames of the entire video is computational expensive. Existing approaches~\cite{wang2016temporal, lei2021less} applied sparse sampling strategies to reduce the computational cost. To sample frames with consistent semantic information, Huang~\etal~\cite{huang2018trailers} proposed dividing the input video into a sequence of coherent shots and randomly select shots as the visual representation. Motivated by this work, we introduce our shot sampling strategy in \Cref{fig:MovieCLIP_framework} (B), with the following steps: 
\smallskip

\noindent\textbf{Shot-based Video Representation.} Given a video $V$, we initially divide it into separate shots $\{S_1, S_2, ... , S_N\}$ using a shot boundary detection framework~\cite{souvcek2020transnet}. Within each shot $S_i$, we uniformly sample $m$ frames $\{I_{i1}, I_{i2}, ..., I_{im}\}$. The representation of shot $S_i$ is computed by taking the average of the features extracted from the $m$ frames:
\begin{align}
    \boldsymbol{\rm f_v}(S_i; \boldsymbol{\theta_v}) = \frac{1}{m} \sum^{m}_{j=1} \boldsymbol{\rm f_v}(I_{ij}; \boldsymbol{\theta_v}),
\end{align}
where $\boldsymbol{\rm f_v}( \cdot )$ is the feature extractor, and $\boldsymbol{\theta}$ its the corresponding parameters. Consistent with our language encoder, we use CLIP as our image encoder $\boldsymbol{\rm f_v}( \cdot )$. We further combine the shot representations to derive the video representation
\begin{align}
    \boldsymbol{\rm f_v}(V; \boldsymbol{\theta_v}) = \frac{1}{N'} \sum^{N'}_{i=1} \boldsymbol{\rm f_v}(S_{i}; \boldsymbol{\theta_v}),
\end{align}
where $N'$ denotes the number of sampled shots. Correspondingly, the number of sampled frames is $T'=m \times N'$, which is much smaller than the number of frames $T$ in the video, thus reducing our model's computational burden. \smallskip

\noindent\textbf{Shot Sampling Strategy.} We observe that in movie videos, shots within the same scene are often semantically similar, resulting redundancy in movie genre classification. For example, as we will show in \Cref{fig:shot_retrieval}, shots within scenes depicting weapons and soldiers always tend to be categorized as the genre of \emph{War}, thus containing redundant information. To address this issue, we sample shots from different scenes instead of the random sampling done in prior work~\cite{huang2018trailers}.
\smallskip

\noindent \textbf{Visual Modality Loss.} Consistent to our language modality loss, we apply binary relevance to train multi-label classifiers for the visual modality. Given the prediction score $\rho_v$, the loss is defined as:
\begin{align}
    L_{vj} &=- \left[ y_{j}\log(\rho_{vj})+(1-y_{j})\log(1-\rho_{vj}) \right],\\
    L_{visual} &= \frac{1}{K}\sum^{K}_{j=1} L_{vj},
\end{align}
where $\rho_{vj}$ denotes the prediction score of visual modality for the $j$-th movie genre.

\subsection{Adaptive Fusion}
\label{sec:movieclip_fusion} 
To combine multimodal features, we apply a weighted linear regression on the outputs of each modality. Let $\rho=\{\rho_v, \rho_l, \rho_a\}$, then the final prediction score is obtained by
\begin{align}
    \rho = \alpha\rho_v+\beta\rho_a+\gamma\rho_l,
\end{align}
where $\alpha$, $\beta$, $\gamma$ can be interpreted as the hyperparameters that control the contribution of each modality. Inspired by adaptive mechanisms that can automatically learn hyper-parameters~\cite{zhang2019adaptive}, we convert $\alpha$, $\beta$, $\gamma$ into extra learnable parameters. Thus, these parameters can be considered as attention weights assigned to each modality. Correspondingly, the loss function of multi-modal features is:
\begin{align}
    L_{j} &=- \left[ y_{j}\log(\rho_{j})+(1-y_{j})\log(1-\rho_{j}) \right],\\
    L_{multi} &= \frac{1}{K}\sum^{K}_{j=1} L_{j},
\end{align}
where $\rho_j$ denotes the prediction scores based on multi-modal features for the $j$-th movie genre. The loss for the whole model is given by:
\begin{align}
    L_{total} = L_{multi}+L_{visual}+L_{text}+L_{audio}.
\end{align}

\section{Experiments}
\label{sec:experiments}

\subsection{Datasets and Experimental Settings}
\label{sec:experiments_settings}

\noindent\textbf{Datasets.} We evaluate Movie-CLIP on MovieNet~\cite{huang2020movienet} and Condensed Movies~\cite{bain2020condensed}. The released version of MovieNet contains 1.1K movies and 30K trailers, which are provided as URLs. Filtering out invalid links and unlabeled trailers resulted in 28,466 trailers remaining. As MovieNet~\cite{huang2020movienet} does not release their testing split, we randomly split the 28K trailers into training, validation, and test sets, following the ratio of 7:1:2 in~\cite{huang2020movienet}. Condensed Movies consists of 33K movie clips from 3,600 movies. After we processed Condensed Movies using the same procedure as MovieNet, we obtained 22,174 movie clips. We split the dataset into 15,521/2,217/4,436 train/val/test clips, respectively. 
\smallskip

\noindent\textbf{Metrics.} 
Following~\cite{huang2020movienet}, we adopt recall@0.5 and precision@0.5, which refers to using a 0.5 threshold to distinguish between positive and negative predictions as our evaluation metrics.  We also use mean average precision (mAP). Given the considerable unbalanced distribution of movie genres, we report the assessment scores at ``macro'' and ``micro'' levels. The ``macro'' average weights the metrics equally for each genre, thereby ignoring the label imbalance. Within the ``micro'' average, the metrics are computed across all categories. It aggregates the contributions of all classes. In simpler terms, ``macro'' accentuates the influence of samples attributed to smaller categories, while ``micro'' assigns equal importance to every single sample.
\smallskip

\noindent\textbf{Baselines.} Following~\cite{huang2020movienet}, we first compare Movie-CLIP with several established models, namely TSN~\cite{wang2016temporal}, I3D~\cite{carreira2017quo}, TRN~\cite{zhou2018temporal}, and Trailer-Storylines~\cite{huang2018trailers}. Our reported results for these models are taken from~\cite{huang2020movienet}. In addition, we extend our evaluation to include adaptations from video classification and retrieval: SlowFast~\cite{feichtenhofer2019slowfast}, Uniformer~\cite{li2023uniformer}, and Imagebind~\cite{girdhar2023imagebind}. We reproduce these models, drawing from either their official code repositories or the MMAction2~\cite{2020mmaction2} framework. To ensure a fair comparison, we finetune these models' video encoders on each dataset and replace their prediction heads with our custom-designed multi-label genre classifiers.
\smallskip

\noindent\textbf{Implementation Details.}
Each input video is split into distinct shots by TransNet v2~\cite{souvcek2020transnet}. We select 8 shots where each shot consists of 3 sampled frames as the visual representation of the input video. We sample audio waveforms at a rate of 16 kHz from each video as the input to both PANNs~\cite{kong2020panns} and Whisper~\cite{radford2023robust}. Our models are trained with a batch size of 256 and a maximum learning rate of $10^{-3}$ on NVIDIA RTX-3090 GPUs. We adopt the      ``ReduceLROnPlateau''~\cite{paszke2019pytorch} strategy to reduce the learning rate.

\begin{table*}[t!]
\centering
\begin{tabular}{l c c c c c c c}

\toprule
\multirow{2}{*}{\emph{Method}}           & \multicolumn{3}{c}{macro} & & \multicolumn{3}{c}{micro} \\
\cline{2-4} \cline{6-8}
\multicolumn{1}{l}{} & r@0.5 & p@0.5 & mAP &  & r@0.5 & p@0.5 & mAP \\ 
\midrule
\multicolumn{8}{c}{\textbf{(A) MovieNet}} \\
\midrule
TSN~\cite{wang2016temporal} & 17.95 & 78.31 & 43.70 & & - & - & - \\
I3D~\cite{carreira2017quo} & 16.54 & 69.58 & 35.79 & & - & - & - \\
TRN~\cite{zhou2018temporal}  & 21.74 & 77.63 & 45.23 & & - & - & - \\
\hline
SlowFast~\cite{feichtenhofer2019slowfast} & 17.60 & 67.70 & 41.33 & & 28.82 & 69.34 & 50.35\\
SlowFast-Multimodal~\cite{xiao2020audiovisual} & 21.05 & 68.12 & 43.62 & & 35.16 & 65.55 & 52.96 \\
Trailer-Storylines~\cite{huang2018trailers}         & 19.52          & 72.40          & 44.02     &    & 33.32          & 64.55          & 53.14  \\
Uniformer~\cite{li2023uniformer} & 38.72 & 70.31 & 58.21 & & 48.61 & 74.19 & 69.00 \\
Imagebind~\cite{girdhar2023imagebind} & 39.84 & 71.63 & 58.74 & & 49.85 & 73.81 & 69.66 \\
Movie-CLIP (ours) & \textbf{40.42} & \textbf{80.05} & \textbf{65.38} & & \textbf{52.45} & \textbf{80.08} & \textbf{75.21}         \\
\midrule
\multicolumn{8}{c}{\textbf{(B) Condensed Movies}} \\
\midrule
SlowFast~\cite{feichtenhofer2019slowfast} & 17.25 & 58.18 & 39.99 & & 26.72 & 65.88 & 50.78\\
SlowFast-Multimodal~\cite{xiao2020audiovisual} & 20.57 & 59.29 & 42.51 & & 34.45 & 67.36 & 53.89 \\

Trailer-Storylines~\cite{huang2018trailers}         & 14.87          & 61.57          & 41.33     &    & 26.39          &  68.95      & 54.83   \\
Uniformer~\cite{li2023uniformer} & 30.25 & 74.92 & 55.58 & & 45.68 & 72.94 & 67.19 \\
Imagebind~\cite{girdhar2023imagebind} & \textbf{42.79} & 76.40 & 63.68 & & 52.42 & 75.58 & 72.73 \\
Movie-CLIP (ours) & 41.30 & \textbf{86.38} & \textbf{72.01} & & \textbf{53.79} & \textbf{82.53} & \textbf{78.35} \\
\bottomrule
\end{tabular}
\vspace{-2mm}
\caption{\textbf{Movie genre classification.} The scores of TSN~\cite{wang2016temporal}, I3D~\cite{carreira2017quo}, TRN~\cite{zhou2018temporal} are cited from~\cite{huang2020movienet}, which do not include the micro metrics. We implement the other baselines and report both macro and micro metrics to provide comprehensive comparison for unbalanced genre labels. Movie-CLIP not only notably outperforms the baselines reported in~\cite{huang2020movienet}, but also achieves a 6$\sim$9\% improvement in mAP compared to methods adapted from different video classification tasks. See \Cref{sec:experiments_genre} for discussion.}
\vspace{-4mm}
\label{table:genre_classification}
\end{table*}

\subsection{Genre Classification}
\label{sec:experiments_genre}

\noindent \textbf{Quantitative Results.}
In \Cref{table:genre_classification}, we present the quantitative results of different models on MovieNet~\cite{huang2020movienet} and Condensed Movies~\cite{bain2020condensed}. We observe that Movie-CLIP significantly outperforms the baselines referenced in~\cite{huang2020movienet}. \Eg, Movie-CLIP outperforms Trailer-Storylines~\cite{huang2018trailers} by approximately 20$\sim$30\% in macro-mAP and 22$\sim$24\% in micro-mAP. The improvement in both macro and micro metrics demonstrates the capability of Movie-CLIP to enhance performance across the entire dataset, including samples within imbalanced genres. When comparing to baselines adapted from other video understanding tasks\footnote{We replace the prediction heads of these models with our multi-label classifiers to make them work better on movie genre classification.}, Movie-CLIP achieves an improvement of 6$\sim$9\% in mAP. Though Movie-CLIP does not achieve the highest performance across all metrics, \eg, Imagebind~\cite{girdhar2023imagebind} achieving the highest r@0.5 score on Condensed Movies, we note that it has a better trade-off among various metrics, thus validating the effectiveness of our proposed modules.
\smallskip

\noindent \textbf{Ablation Study.} In \Cref{table:ablation_study}, we present the ablation results of Movie-CLIP from three aspects. First, we compare our shot sampling strategy (\emph{Shot}) to randomly sampling (\emph{Random}). Our shot sampling strategy notably outperforms randomly sampling. Second, we evaluate the effectiveness of each modality. We note that while the performance does not match that of the \emph{Shot} modality (\eg, 62.2 vs.\ 42.5 vs.\ 37.8 in macro-mAP), the \emph{Audio} and \emph{Language} modalities still hold significant importance in movie genre prediction. Third, we investigate the incorporation of text features and observe that incorporating text features improves performance compared to models lacking such features (\eg, 64.8 vs.\ 65.2 vs.\ 65.4 in macro-mAP, 75.0 vs.\ 74.8 vs.\ 75.2 in micro-mAP ). The improvement validates that our language augmentation module can effectively extract pertinent linguistic information from the input audio, while also verifying our keyword-aware documents can suppress the noise in the initial transcript.



\begin{table}[t]
\centering
\begin{tabular}{l c c}
\toprule
\emph{Method} & macro-mAP & micro-mAP \\
\midrule
Random & 52.81 & 63.90 \\
Shot & 62.22 & 72.31 \\
Audio & 42.48 & 58.92 \\
Language & 37.76 & 49.04 \\
Shot+Audio & 64.77 & 75.00 \\
Shot+Audio+Raw Text & 65.15 & 74.80 \\
Shot+Audio+Language & \textbf{65.38} & \textbf{75.21} \\
\bottomrule
\\
\end{tabular}
\vspace{-2mm}
\caption{\textbf{Ablation study on MovieNet.} \emph{Random} denotes that we randomly sample frames from the videos. \emph{Shot} denotes our shot sampling strategy. \emph{Raw Text} denotes that we directly use the initial transcript $L$ as the text input. \emph{Language} denotes that we use the keyword-aware document as the text input. }
\vspace{-4mm}
\label{table:ablation_study}
\end{table}

\subsection{Movie Analysis based on Genre Classification}
\label{sec:experiments_analysis}
\noindent \textbf{Genre-based Shot Retrieval.} As discussed in the Introduction, comprehending movies is a challenging task, especially when dealing with videos of considerable duration. However, given that Movie-CLIP segments input videos into discrete shots, it can be easily applied to genre-based shot retrieval in lengthy videos through a sliding window approach. Correspondingly, Movie-CLIP extracts a sequence of genre labels from the video input.

\begin{figure*}[t]
    \centering
    \includegraphics[width=.9\linewidth]{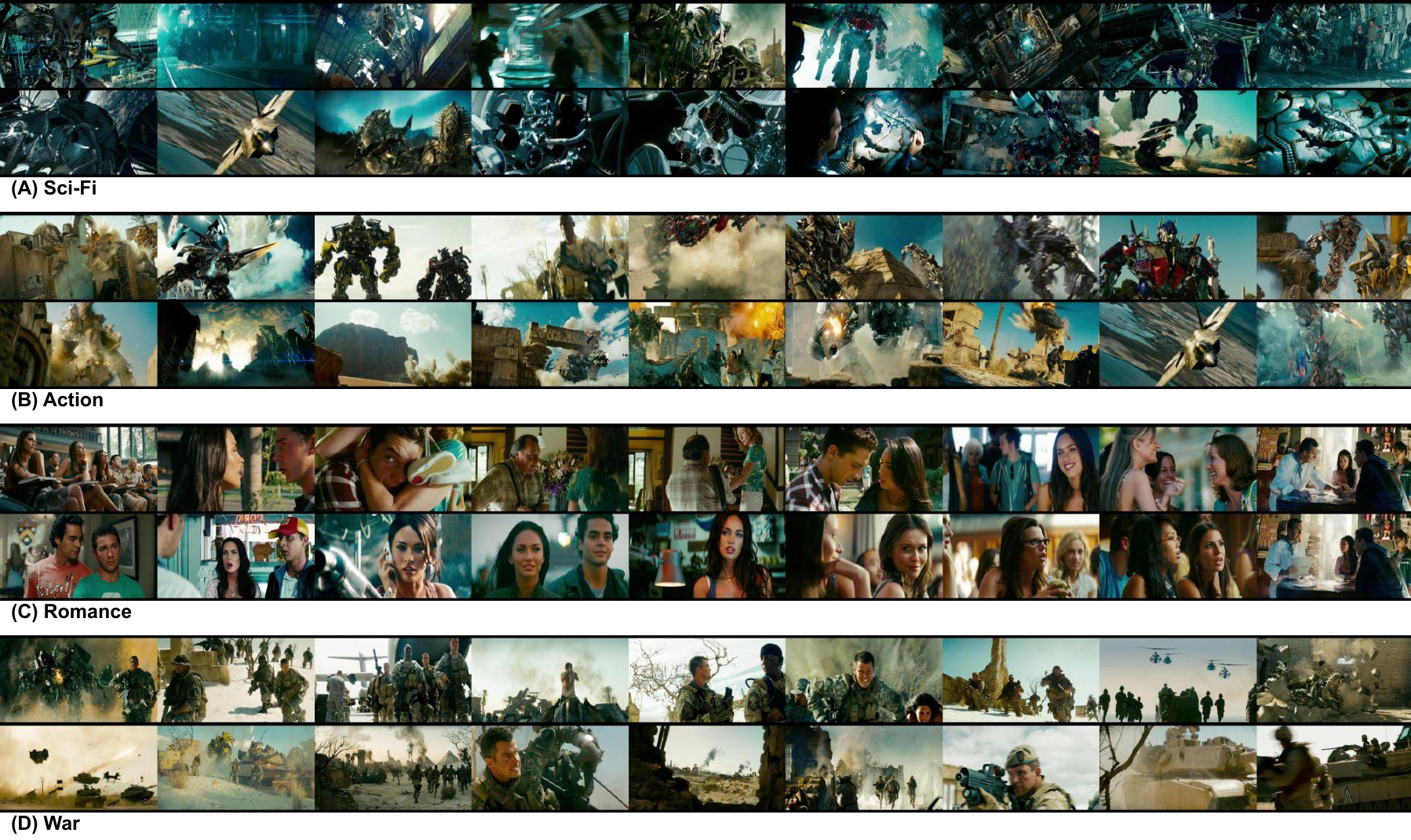}
    \vspace{-2mm}
    \caption{\textbf{Genre-based shot retrieval} on movie ``Transformers: Revenge of the Fallen.'' Movie-CLIP effectively identified shots corresponding to various genres within a video spanning hours. See Section \ref{sec:experiments_analysis} for detailed discussion}
    \vspace{-4mm}
    \label{fig:shot_retrieval}
\end{figure*}

In \Cref{fig:shot_retrieval}, we present the application of genre classification in genre-based shot retrieval using ``Transformers: Revenge of the Fallen'' as an example. As shown in the figure, Movie-CLIP not only accurately identifies shots that correspond to the ground truth genres but also generalizes well to genres that are not part of the ground truth. Specifically, the movie ``Transformers: Revenge of the Fallen'' falls under the genres of \emph{Sci-Fi} and \emph{Action}, with their respective shots showcased in \Cref{fig:shot_retrieval} (A) and \Cref{fig:shot_retrieval} (B). Aligning with our expectations, shots classified as \emph{Sci-Fi} encompass scenes depicting the universe, planets, or robot armies, while shots categorized as \emph{action} show up together with typical elements found in \emph{action} movies, such as explosions or dynamic movements. We further present two additional genres, \emph{Romance} and \emph{War}, in \Cref{fig:shot_retrieval} (C) and \Cref{fig:shot_retrieval} (D), respectively. Movie-CLIP effectively shows a sequence of shots related to these two genres. For example, in \Cref{fig:shot_retrieval} (D), shots featuring weapons or soldiers are more likely to associated with the \emph{War} genre, whereas shots portraying daily life or romantic relationships are more likely to align with the \emph{Romance} genre. Genre-based shot retrieval carriers practical implications, such as automated trailer generation or automatic clipping for highlighting movies. See the supplementary for additional examples.
\smallskip

\begin{figure*}[t]
    \centering
    \includegraphics[width=\linewidth]{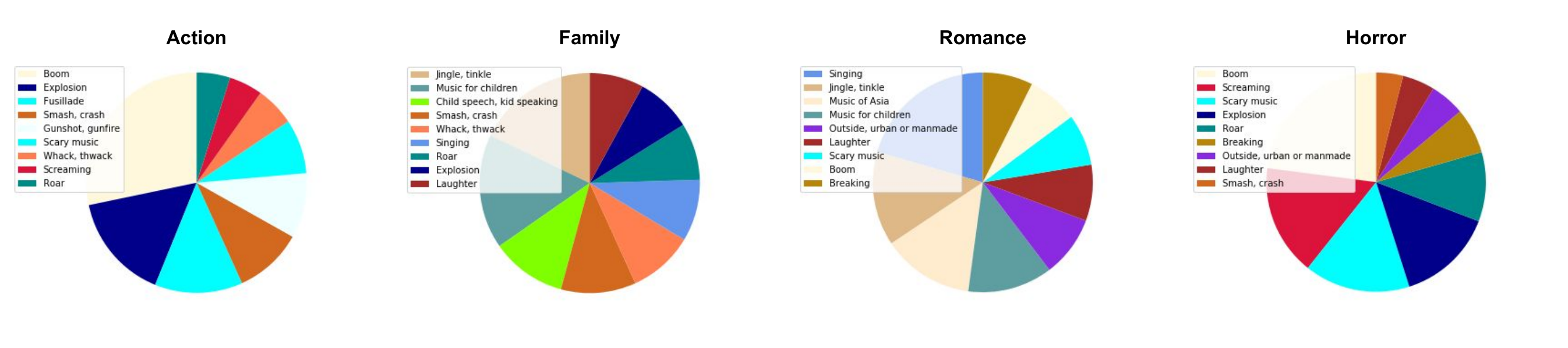}
    \vspace{-2mm}
    \caption{\textbf{Representative sound events} of \emph{Action}, \emph{Family}, \emph{Romance} and \emph{Horror}. Sound events exhibit discriminative characteristics acorss various genres, supporting our motivation for using sound events to improve our model. See \Cref{sec:experiments_analysis} for further discussion.}
    \vspace{-4mm}
    \label{fig:sound_event}
\end{figure*}

\begin{figure}[t]
    \centering
    \includegraphics[width=\linewidth]{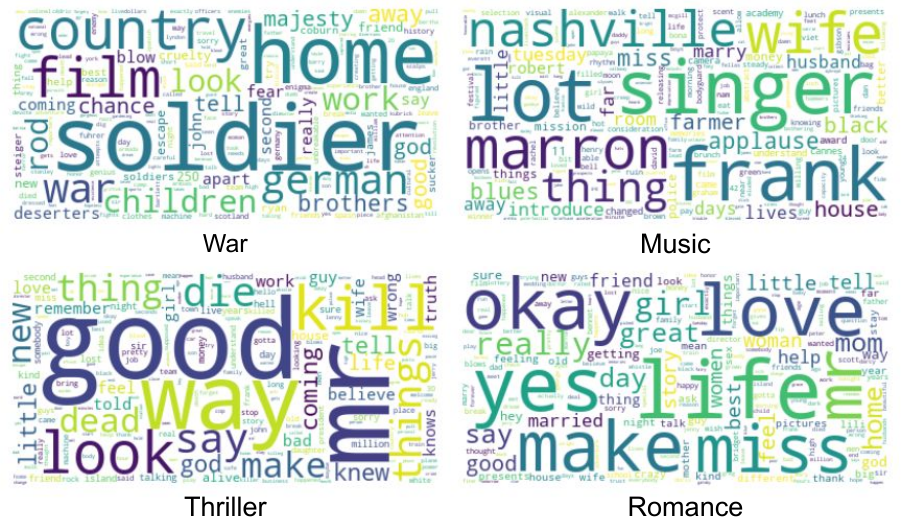}
    \caption{\textbf{Wordclouds} of \emph{War}, \emph{Music}, \emph{Thriller} and \emph{Romance} (MovieNet). Word clouds show varying distributions across genres, validating our keyword-aware approach to movie genre classification. See Section \ref{sec:experiments_analysis} for further discussion.}
    \label{fig:wordclouds}
\end{figure}

\noindent \textbf{Sound Event Analysis.} We analyze audio waveforms on MovieNet to uncover the correlations between genres and sound events. The representative sound events of four different movie genres are presented in \Cref{fig:sound_event}. The figure substantiates that sound events in the audio modality are discriminative attributes for genre recognition. For example, the prevalent sound events in the \emph{Romance} genre include ``Singing,'' ``Music for children,'' and ``jingle, tinkle,'' evoking feelings of relaxation and happiness. In contrast, elements characteristic of \emph{Action} movies consistently involve ``Gunshot,'' ``Scary music,'' and ``fusillade,'' making people feel thrilled and excited. Additional examples can be found in the supplementary material.
\smallskip

\noindent \textbf{Keyword Analysis.}
We calculate the Term Frequency - Inverse Document Frequency (TF-IDF) to reveal the correlations between keywords and movie genres. Specifically, for each genre, we create table $T$ with dimensions $n \times m$, where $n$ is the number of movies in that genre and $m$ represents the vocabulary size. $T_{ij}$ denotes the TF-IDF value of word $j$ in movie $i$, and the score of word $j$ across entire genre is defined as $s_j = \sum_i^n T_{ij}$. Subsequently, we generate wordclouds for each genre. It is worth noting that some words such as ``know,'' ``man,'' ``think'' rank high among most of genres but do not carry real information. To address this issue, we introduce a mechanism where the top $N$ words from all genres are aggregated into a list of size $N\times21$, and their occurrences are counted. Any word surpassing the threshold of $M$ occurrences within this list is excluded from the wordcloud plots. Here, we set $N$ to 20 and $M$ to 5. \Cref{fig:wordclouds} shows wordcloud plots of \emph{Romance}, \emph{Thriller}, \emph{Music} and \emph{War} genres on the trailer data of MovieNet. We observe that trailers containing keywords such as ``singer,'' ``applause,'' ``blues'' tend to align more with the \emph{Music} genre. Conversely, \emph{War} movies exhibit stronger associations with words such as ``soldier,'' ``country,'' ``majesty.'' See supplementary for additional examples.

\subsection{Generalization to Scene Boundary Detection}
\label{sec: boundary detection}
\noindent \textbf{Datasets \& Experiment Settings.} The evaluation of the scene boundary detection task is conducted on MovieNet using 318 movies with annotated scene boundaries. Following ShotCoL~\cite{chen2021shot}, we split the 318 movies into 190, 64, 64 movies for training, validation and test sets, respectively. Average Precision (AP) and Recall@0.5 are applied as our evaluation metrics. For this task, we use a sequence of four consecutive shots as input, with the probability that a scene boundary exists between the second and third shots as output. Consistent with our approach in the genre classification task, we employ Binary Cross Entropy as the loss function. Due to the data imbalance, the weight for boundary versus non-boundary samples is 10:1.
\smallskip

\noindent \textbf{Model Architecture.} Following~\cite{chen2021shot}, our decoder uses a three-layer MLP classifier (number-of-shots $\times$ feature-dimension - 4096 - 1024 -2 ). Similar to the genre classification task, we representation each shot with 3 subsampled frames. As in~\cite{chen2021shot}, we encode each shot's frames using a model pretrained on Places\footnote{Places is a large-scale dataset for the scene recognition task.}~\cite{zhou2017places}.
\smallskip

\noindent \textbf{Evaluation Results.} Table \ref{tab: scene_boundary} reports the scene boundary detection results of Movie-CLIP and other baselines. From the table, we see that Movie-CLIP gets new state-of-the-art results with our shot representations and Places features.



\begin{table}[t]
\centering
\begin{tabular}{ l c c}
\toprule
 Models & AP & Recall@0.5 \\
\midrule
SCSA~\cite{chasanis2008scene} & 14.7 & 54.9 \\
Story Graph~\cite{tapaswi2014storygraphs}&25.1 & 58.4\\
Siamese~\cite{baraldi2015deep} &28.1 & 60.1 \\
ImageNet~\cite{deng2009imagenet} & 41.26 & 30.06 \\
Places~\cite{zhou2017places} &  43.23 & 59.34\\
LGSS~\cite{rao2020local} &  47.1 & 73.6 \\
ShotCoL~\cite{chen2021shot} & 53.37 & 81.33 \\
Movie-CLIP (ours) & \textbf{54.45} & \textbf{82.21} \\
\bottomrule \\
\end{tabular}
\caption{\textbf{Scene boundary detection} results on MovieNet. Baseline results are taken from \cite{chen2021shot}. We find Movie-CLIP generalizes well on the scene boundary detection task, outperforming the state-of-the-art.  See Section \ref{sec: boundary detection} for discussion.}
\label{tab: scene_boundary}
\end{table}

\subsection{Limitations and Future Work}
\label{sec: limitations}
In this paper, we adopt the binary relevance strategy to train Movie-CLIP. While this approach is straightforward to implement, it ignores the inter-dependencies among labels. Our observation reveals that movie genres indeed exhibit correlations with one another. For example, genres like \emph{Thriller}, \emph{Crime}, and \emph{Horror} often co-occur. Similarily, \emph{Family} genre frequently occurs with \emph{Animation}. In contrast, negative Pearson correlation coefficients exist between genres like \emph{Comedy} and \emph{Thriller}, \emph{Drama} and \emph{Documentary}. Based on these findings, we conclude that effectively leveraging the correlations among different genres should be helpful for movie genre classification.

Moreover, our primary focus lies in the impact of pretrained features from different modalities in this paper. The encoders to extract these features remain frozen during the training of Movie-CLIP. As a result, a potential improvement can involve the development of refined training strategies, such as end-to-end learning method.

\section{Conclusion}
In this paper, we propose a movie genre classification model, Movie-CLIP, which consists of language augmentation and shot sampling modules. For language augmentation model, since our model's transcripts are extracted from input audios, Movie-CLIP enhances performance without requiring additional language annotations. Additionally, we introduce a shot sampling strategy designed to select representative shots from diverse scenes within a video. This approach reduces computational cost in comparison to encoding the entire video. Movie-CLIP outperforms existing benchmarks on movie genre classification, improving 6-9\% mAP scores on MovieNet and Condensed Movies datasets. We also generalize Movie-CLIP to scene boundary detection task, achieving the new state-of-the-art by improving 1.1\% AP scores. We perform extensive experiments to demonstrate the applications of Movie-CLIP on movie analysis and explore the correlations between genres and various movie elements across different modalities.

\noindent\textbf{Acknowledgements}
This material is based upon work supported, in part, by DARPA under agreement number HR00112020054. Any opinions, findings, and conclusions or recommendations are those of the author(s) and do not necessarily reflect the views of the supporting agencies.


{\small
\bibliographystyle{ieee_fullname}
\bibliography{egbib}
}

\appendix

\section{Expanded Experimental Results}
\subsection{Per-genre performance}
\label{experiments_per_genre_performance}
\begin{figure*}[t]
    \centering
    \includegraphics[width=0.8\linewidth]{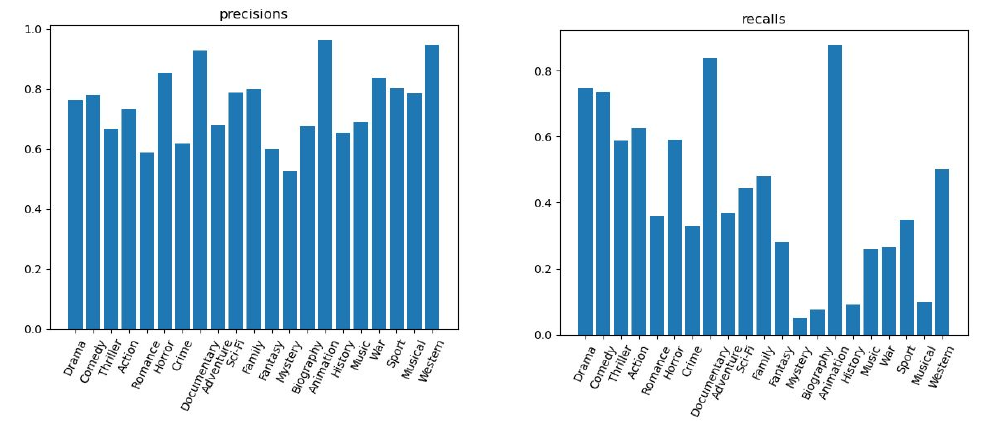}
     \caption{Per-genre performance of Movie-CLIP on MovieNet (Left: precision; Right: recall). See \Cref{experiments_per_genre_performance} for discussion.}
    \label{fig:per-genre_performance}
\end{figure*}

In \Cref{fig:per-genre_performance}, we present the performance of Movie-CLIP across each genre on the MovieNet dataset. We observe that Movie-CLIP achieves robust performance across a majority of genres. However, the accurate prediction of genres such as \emph{Mystery}, \emph{Biography}, or \emph{History} is still challenging. This difficulty may arise due to the imbalanced label distribution among various genres. Therefore, exploring methods to solve the long-tail distribution issue can be promising to further improve our model's performance. 

\subsection{Low-level Visual Attribute Analysis.} 
\label{experiments_lowlevel_visual}
\begin{figure*}[t]
    \centering
    \includegraphics[width=1.0\linewidth]{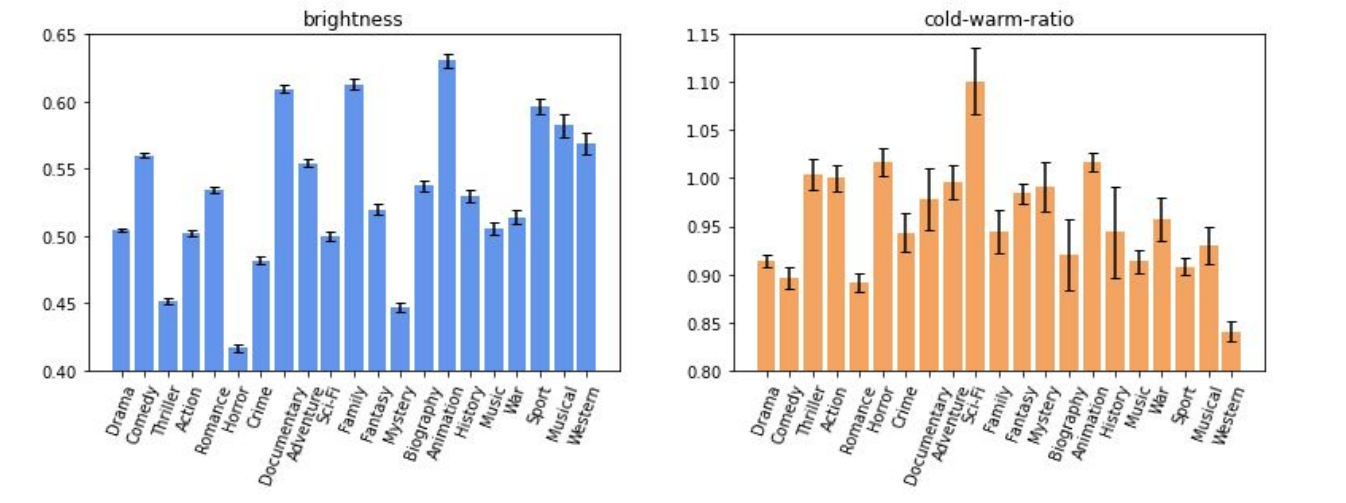}
    \caption{Low-level visual feature analysis across movie genres. Left: brightness with confidence interval; Right: cold-warm color ratio with confidence interval. See \Cref{experiments_lowlevel_visual} for discussion.}
    \label{fig:low_level_visual}
\end{figure*}

In \Cref{fig:low_level_visual}, we analyze the distribution of brightness and warm-cold color ratio across genres to explore correlations between genres and low-level visual attributes. From the figure, we observe that \emph{horror} genre exhibits the lowest brightness value, which is consistent with the intuition that \emph{horror} films aim to evoke the fear emotion through dark and ominous atmospheres. In contrast, genres such as \emph{family} and \emph{animation} tend to favor higher brightness values. This trend corresponds with the inherent intent of these genres, which is to convey feelings of love and warmth to their audience.

Regarding the cold-warm color ratio, \emph{western} films demonstrate the lowest values, while \emph{Sci-Fi} achieves the highest values. This contrast can be attributed to the characteristics of these genres, \ie, \emph{Western} films often has a sepia tone due to scenes like desert, blazing sun, and dirt. On the other hand, \emph{Sci-Fi} movies lean towards colder colors for scenes such as the universe, spacecraft, or robot armies, contributing to an impression of high-tech and sharpness

\subsection{Ablation Study on Visual Features}
\label{experiments_ablation_study}
To evaluate the effectiveness of our proposed mechanism when using various pretrained visual features, we present the experiment results where we replaced CLIP features with ResNet-50 in \Cref{table:additional_ablation_study}. We observe that our method continues to improve the performance compared to the base models (44.16 vs. 47.40 vs. 49.25 in macro-mAP), validating that the effectiveness of our method does not simply rely on CLIP features.

\begin{table}[t]
\centering
\begin{tabular}{l c c}
\toprule
\emph{Method} & macro-mAP & micro-mAP \\
\midrule
Shot (ResNet-50) & 44.16 & 54.29 \\
Shot+Audio & 47.40 & 57.16 \\
Shot+Audio+Language & 49.25 & 62.53 \\
\bottomrule
\\
\end{tabular}
\caption{\textbf{Applying ResNet-50 features as visual representations on MovieNet:} We still observe performance improvements by our proposed approach, even the visual representations are based on ResNet-50 features.} 
\label{table:additional_ablation_study}
\end{table}

\subsection{Additional Sound Event Analysis}
\label{experiments_sound_event}
\begin{figure*}[t]
    \centering
    \includegraphics[width=.9\linewidth]{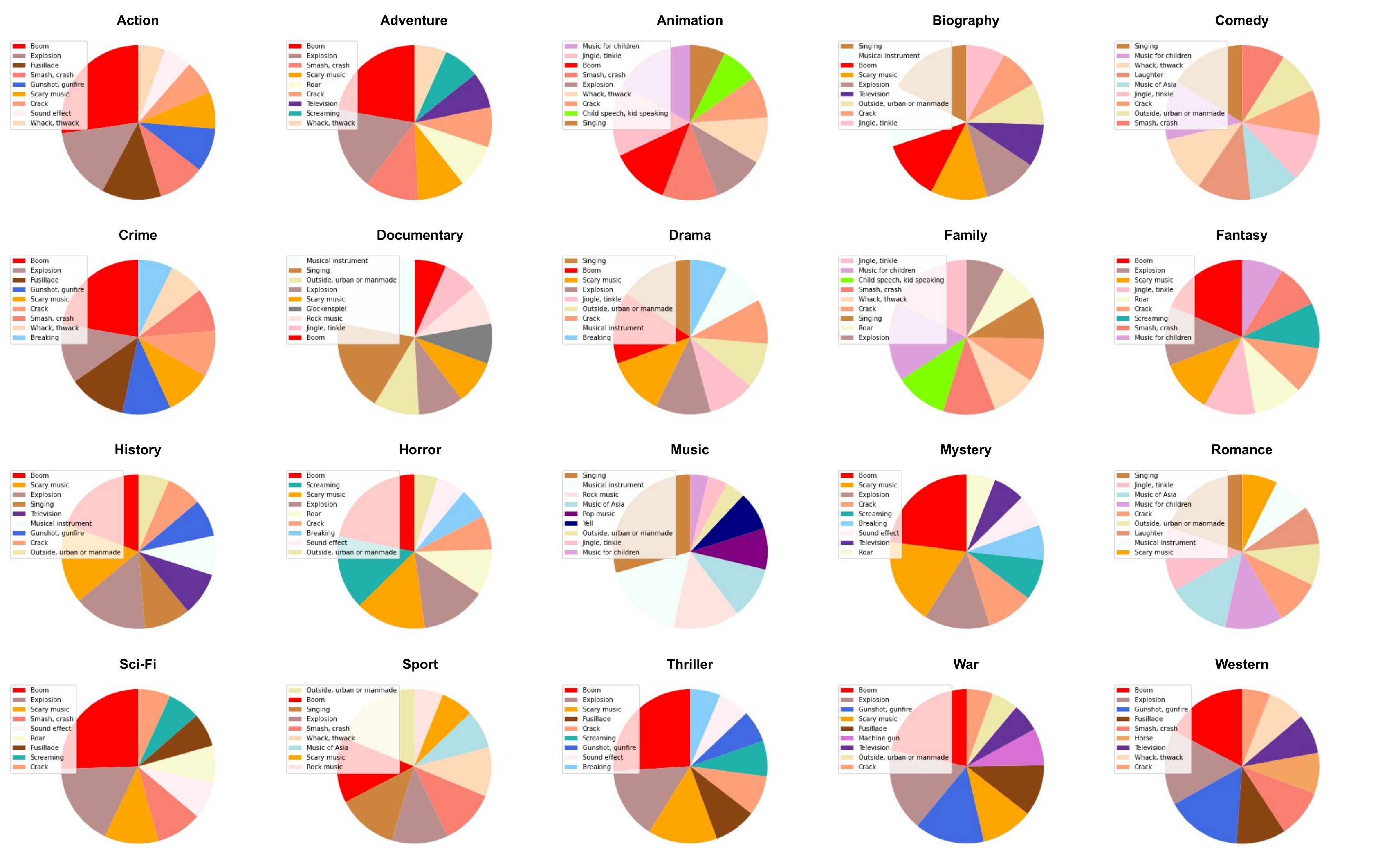}
     \caption{Sound events of different movie genres on MovieNet. See \Cref{experiments_sound_event} for discussion.}
     \vspace{-6mm}
    \label{fig:soundevent_full}
\end{figure*}

\Cref{fig:soundevent_full} provides sound events of various movie genres, supplementing the results from the main paper. The distinctive attributes of these sound events substantiate our motivation for incorporating them to augment our model.

\subsection{Additional Keyword Analysis}
\label{experiments_keyword}
\begin{figure*}[t]
    \centering
    \includegraphics[width=.9\linewidth]{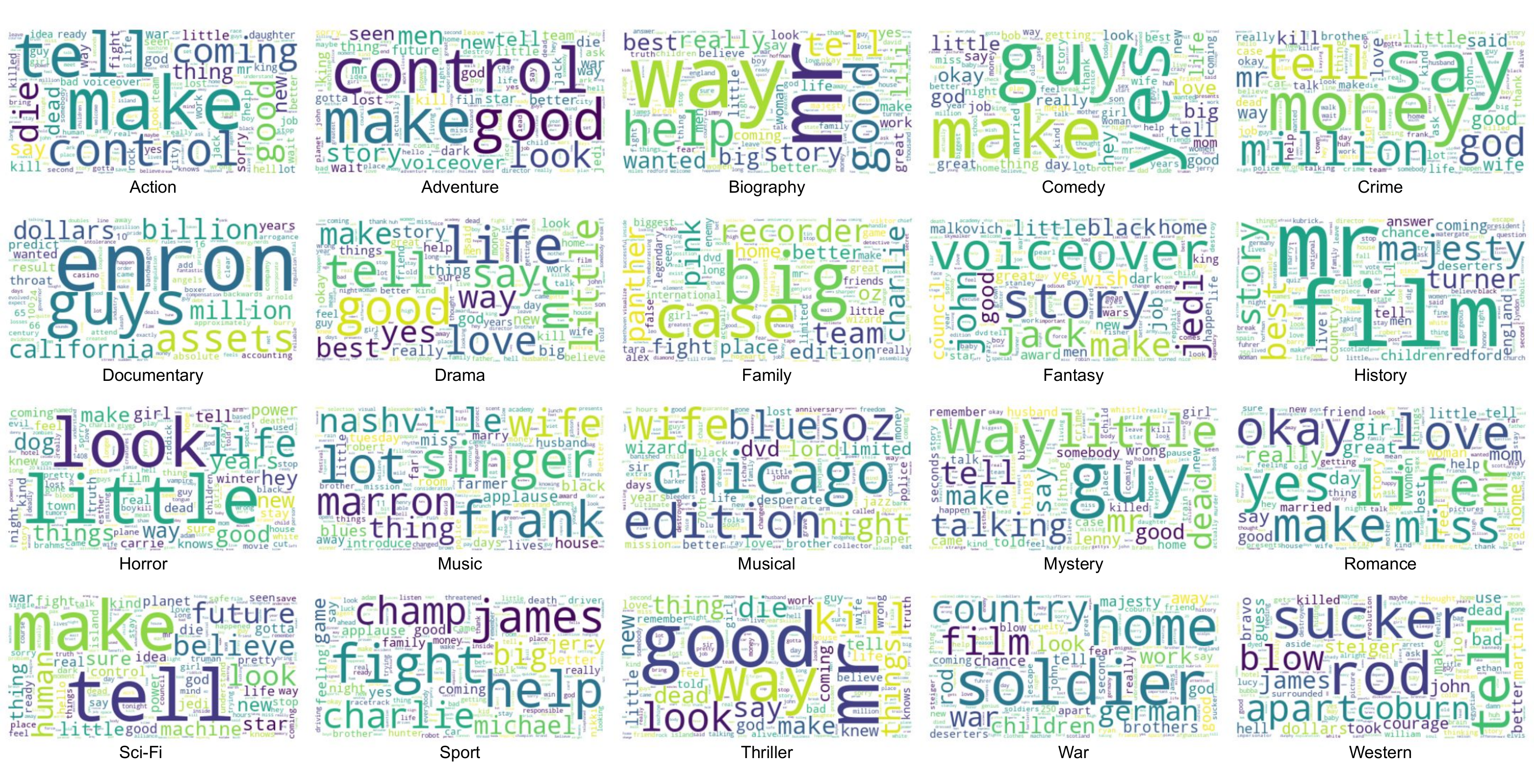}
    \caption{Wordclouds of different movie genres on MovieNet. See \Cref{experiments_keyword} for discussion.}
     \vspace{-6mm}
    \label{fig:wordcloud_full}
\end{figure*}

We plot wordclouds in \Cref{fig:wordcloud_full} to supplement the main paper. These visualizations provide additional support for using keyword-aware documents to improve our model.

\subsection{Additional Genre-based Shot Retrieval}
\label{experiments_shot_retrieval}
\begin{figure*}[t]
    \centering
    \includegraphics[width=0.99\linewidth]{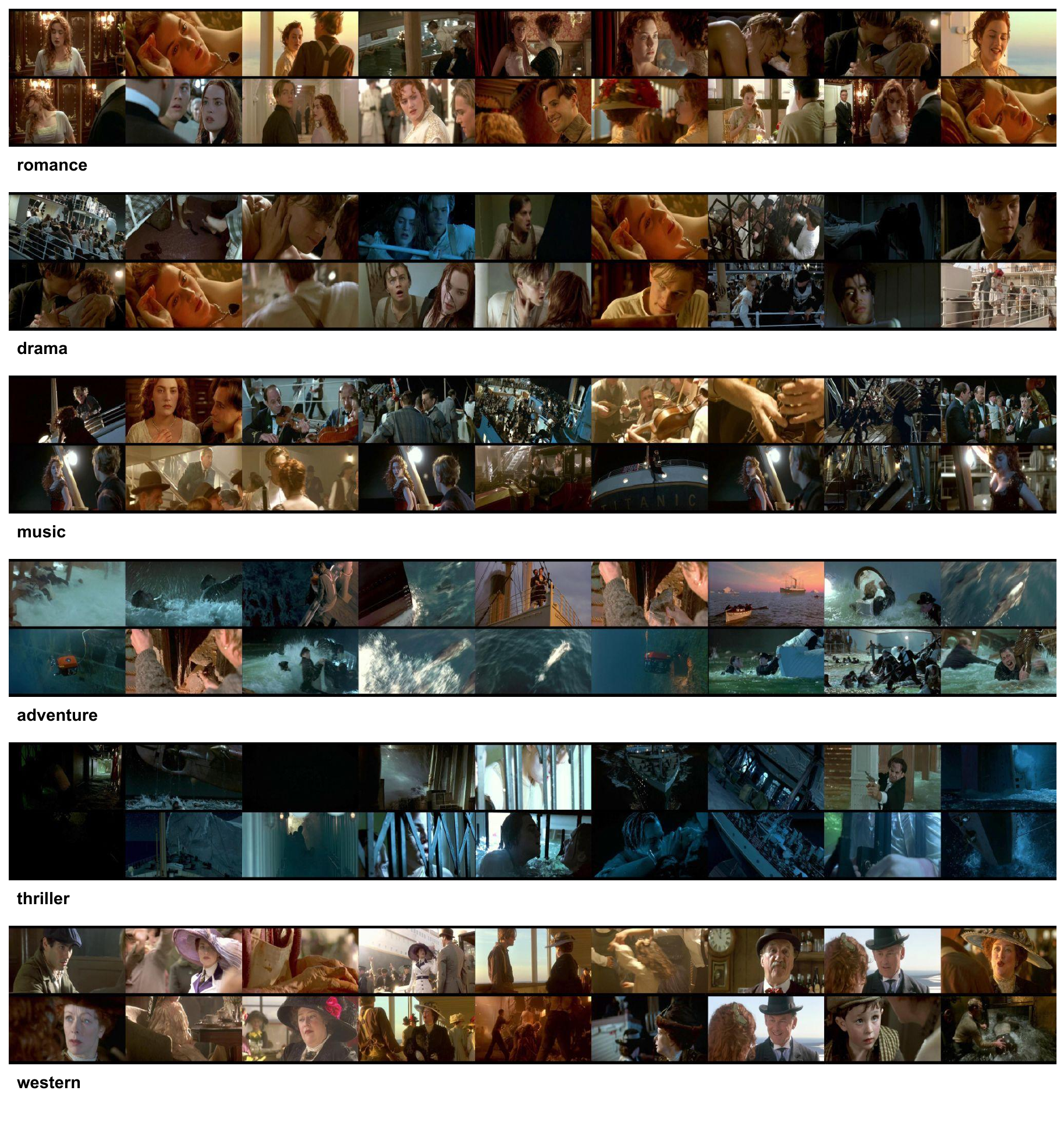}
     \caption{Genre-based shot retrieval on ``Titanic.'' See \Cref{experiments_shot_retrieval} for discussion.}
    \label{fig:shot_retrieval_titanic}
\end{figure*}

\begin{figure*}[t]
    \centering
    \includegraphics[width=0.99\linewidth]{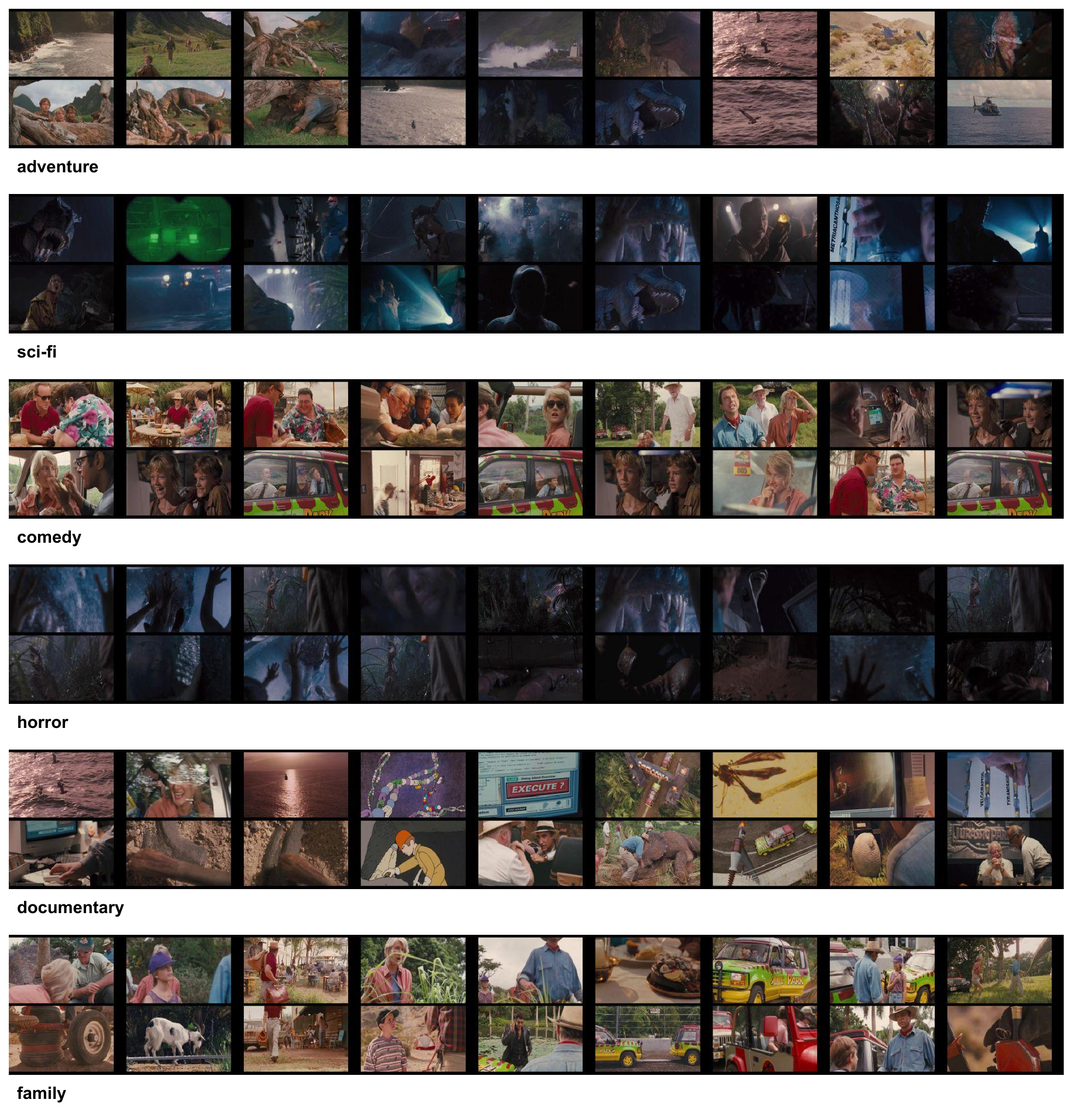}
     \caption{Genre-based shot retrieval on ``Jurassic Park.'' See \Cref{experiments_shot_retrieval} for discussion.}
    \label{fig:shot_retrieval_jurassic}
\end{figure*}

Additional examples of genre-based shot retrieval, Titanic (1997) and Jurassic Park (1993), are presented in \Cref{fig:shot_retrieval_titanic} and \Cref{fig:shot_retrieval_jurassic}, demonstrating that Movie-CLIP effectively identified shots aligned with different genres across video content of hours duration.

\section{Statistics of Videos}
\label{experiments_statistics}
\begin{figure*}[t]
    \centering
    \includegraphics[width=0.8\linewidth]{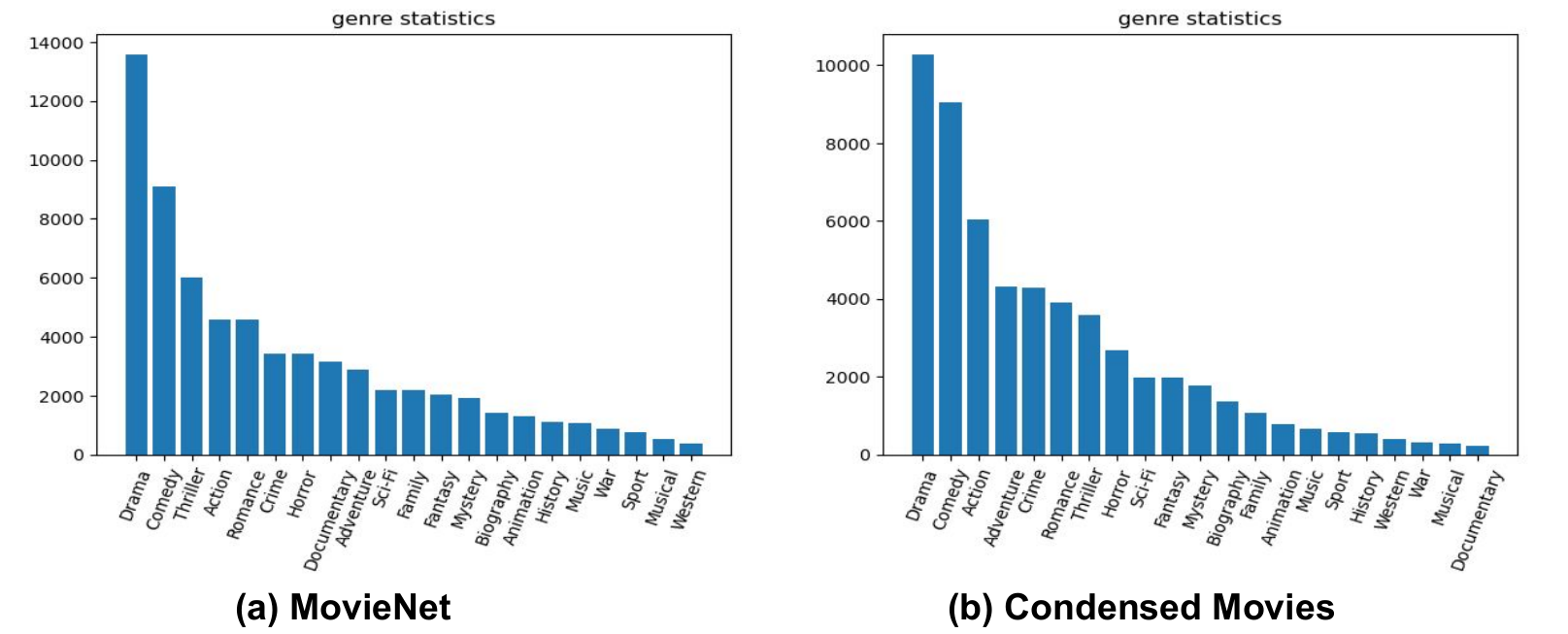}
     \caption{Distribution of genres on MovieNet (left) and Condensed Movies (right). See \Cref{experiments_statistics} for discussion.}
    \label{fig:genre_statistics}
\end{figure*}

We present the statistics of source videos that we used in MovieNet and Condensed Movies in \Cref{tab:dataset_statistics} and \Cref{fig:genre_statistics}. From \Cref{fig:genre_statistics}, we observe that the label distribution is remarkably imbalanced, validating the significance of using ``micro'' and ``macro'' metrics in our experiments.

\begin{table}[t]
    \centering
    \begin{tabular}{l|c c}
        \toprule
         Datasets & MovieNet & Condensed Movies\\
         \midrule
         Type of Video & Trailer & Movie Clip \\
         Total & 28,466 & 22,174 \\
         Training & 19,926 & 15,521 \\
         Validation & 2,846 & 2,217 \\
         Test & 5,694 & 4,436 \\
         \bottomrule
    \end{tabular}
    \caption{Statistics of source videos on MovieNet and Condensed Movies.}
    \label{tab:dataset_statistics}
\end{table}

\end{document}